\RequirePackage[l2tabu, orthodox]{nag}
\documentclass[11pt]{article}

\usepackage[T1]{fontenc}

\usepackage[bitstream-charter]{mathdesign}
\usepackage{amsmath}
\usepackage[scaled=0.92]{PTSans}

\usepackage{paralist}

\usepackage[
  paper  = letterpaper,
  left   = 1.65in,
  right  = 1.65in,
  top    = 1.0in,
  bottom = 1.0in,
  ]{geometry}

\usepackage[usenames,dvipsnames]{xcolor}
\definecolor{shadecolor}{gray}{0.9}

\usepackage[final,expansion=alltext]{microtype}
\usepackage[english]{babel}
\usepackage[parfill]{parskip}
\usepackage{afterpage}
\usepackage{framed}

\DeclareRobustCommand{\parhead}[1]{\textbf{#1}~}

\usepackage{lineno}

\usepackage{ragged2e}

\newcounter{parcount}

\usepackage{graphicx}
\usepackage{natbib}
\usepackage[labelfont=bf]{caption}

\usepackage{booktabs}
\usepackage{multirow}

\usepackage[algoruled]{algorithm2e}
\usepackage{listings}
\usepackage{fancyvrb}
\fvset{fontsize=\normalsize}

\usepackage[colorlinks,linktoc=all]{hyperref}
\usepackage[all]{hypcap}
\hypersetup{citecolor=MidnightBlue}
\hypersetup{linkcolor=black}
\hypersetup{urlcolor=MidnightBlue}

\usepackage[acronym,smallcaps,nowarn]{glossaries}

\lstdefinestyle{mystyle}{
    commentstyle=\color{OliveGreen},
    keywordstyle=\color{BurntOrange},
    numberstyle=\tiny\color{black!60},
    stringstyle=\color{MidnightBlue},
    basicstyle=\ttfamily,
    breakatwhitespace=false,
    breaklines=true,
    captionpos=b,
    keepspaces=true,
    numbers=left,
    numbersep=5pt,
    showspaces=false,
    showstringspaces=false,
    showtabs=false,
    tabsize=2
}
\usepackage[colorinlistoftodos,
           prependcaption,
           textsize=small,
           backgroundcolor=yellow,
           linecolor=lightgray,
           bordercolor=lightgray]{todonotes}

\DeclareRobustCommand{\E}[2]{\mathbb{E}_{#1}\left[#2\right]}

\newcommand{\g}{\, | \,}

\newcommand{\Lcal}{\mathcal{L}}

\newcommand{\bw}{\mathbf{w}}

 \newacronym{ADVI}{advi}{automatic differentiation variational inference}
\newacronym{AR}{a{\small\&}r}{augment and reduce}

\newacronym{BBVI}{bbvi}{black-box variational inference}

\newacronym{CBOW}{cbow}{continuous bag-of-words}
\newacronym{CDF}{cdf}{cumulative distribution function}
\newacronym{CS-EFE}{cs-efe}{context selection for exponential family embeddings}
\newacronym{CTM}{ctm}{correlated topic model}

\newacronym[\glslongpluralkey={deep exponential families}]{DEF}{def}{deep exponential family}
\newacronym{DMIS}{dmis}{deterministic multiple importance sampling}
\newacronym{DLDA}{d-lda}{dynamic latent {D}irichlet allocation}

\newacronym{EFE}{efe}{exponential family embeddings}
\newacronym{ELBO}{elbo}{evidence lower bound}
\newacronym{EM}{em}{expectation maximization}
\newacronym{ETM}{etm}{embedded topic model}
\newacronym{DETM}{d-etm}{dynamic embedded topic model}

\newacronym{GNTS}{gn-ts}{gamma-normal time series model}
\newacronym{G-REP}{g-rep}{generalized reparameterization}

\newacronym{HMC}{hmc}{{H}amiltonian {M}onte {C}arlo}

\newacronym{KL}{kl}{{K}ullback-{L}eibler}

\newacronym{LDA}{lda}{latent {D}irichlet allocation}
\newacronym{LSTM}{lstm}{long short-term memory}

\newacronym{MAP}{map}{\emph{maximum a posteriori}}
\newacronym{MCMC}{mcmc}{{M}arkov chain {M}onte {C}arlo}
\newacronym{MF}{mf}{matrix factorization}
\newacronym{MIS}{mis}{multiple importance sampling}

\newacronym{NVDM}{nvdm}{neural variational document model}

\newacronym{OBBVI}{o-bbvi}{overdispersed black-box variational inference}
\newacronym{OVE}{ove}{one-vs-each}

\newacronym{SIVI}{sivi}{semi-implicit variational inference}
\newacronym{SVI}{svi}{stochastic variational inference}

\newacronym{TMES}{tmes}{topic model in embedding space}

\newacronym{USIVI}{uivi}{unbiased implicit variational inference}
\newacronym{UN}{un}{{U}nited {N}ations}

\newacronym{VAE}{vae}{variational autoencoder}
\newacronym{VEM}{vem}{variational expectation maximization}
\newacronym{VI}{vi}{variational inference}

\usepackage{subfigure}
\usepackage{authblk}
\usepackage{enumitem} 
\usepackage{bm}

\title{\textbf{The Dynamic Embedded Topic Model}}

\author[1, $\ast$]{Adji B. Dieng}
\author[2, 3, $\ast$]{Francisco J. R. Ruiz}
\author[1, 2]{\\David M. Blei}
\affil[1]{Department of Statistics, Columbia University}
\affil[2]{Department of Computer Science, Columbia University}
\affil[3]{Department of Engineering, University of Cambridge}
\affil[$\ast$]{Equal Contributions}

\begin{document}
\maketitle

\begin{abstract}
\noindent Topic modeling analyzes documents to learn meaningful patterns of
  words. For documents collected in sequence, dynamic topic models
  capture how these patterns vary over time. We develop the
  \gls{DETM}, a generative model of documents that combines \gls{DLDA}
  and word embeddings. The \gls{DETM} models each word with a
  categorical distribution parameterized by the inner product between
  the word embedding and a per-time-step embedding representation of
  its assigned topic. The \gls{DETM} learns smooth topic trajectories
  by defining a random walk prior over the embedding representations
  of the topics. We fit the \gls{DETM} using structured amortized
  variational inference with a recurrent neural network. On three
  different corpora---a collection of United Nations debates, a set of
  ACL abstracts, and a dataset of Science Magazine articles---we found
  that the \gls{DETM} outperforms \gls{DLDA} on a document completion
  task. We further found that the \gls{DETM} learns more diverse and
  coherent topics than \gls{DLDA} while requiring significantly less
  time to fit.\footnote{\textbf{Code:} The code for this paper can be found at \url{https://github.com/adjidieng/DETM}}
\end{abstract}

\section{Introduction}
\label{sec:introduction}
\glsresetall

Topic models are useful tools for the statistical analysis of document
collections \citep{blei2003latent,blei2012probabilistic}. They have
been applied to documents from many fields, including marketing,
sociology, political science, and the digital humanities; see
\citet{boydgraber2017applications} for a review.  One of the most
common topic models is \gls{LDA} \citep{blei2003latent}, a
probabilistic model that represents each topic as a distribution over
words and each document as a mixture of the topics.  \gls{LDA} has
been extended in different ways, for example to capture correlations
among the topics \citep{Lafferty2005correlated}, to classify documents
\citep{blei2007supervised}, or to analyze documents in different
languages \citep{Mimno2009polylingual}.

In this paper, we focus on analyzing the temporal evolution of topics
in large document collections. Given a corpus that was collected over
a large number of years, our goal is to use topic modeling to find how
the latent patterns of the documents change over time.

\Gls{DLDA} \citep{blei2006dynamic} shares the same goal. \gls{DLDA} is
an extension of \gls{LDA} that uses a probabilistic time series to
allow the topics to vary smoothly over
time.\footnote{\citet{blei2006dynamic} called it a \emph{dynamic topic
    model}, but we refer to it as \gls{DLDA} because it is motivated
  as a dynamic extension of \gls{LDA}.}  However, \gls{DLDA} suffers
from the same limitations as \gls{LDA}. In particular, it does not
capture the distribution of rare words and the long tail of language
data \citep{dieng2019topic}.

The \gls{ETM} aims to solve these problems \citep{dieng2019topic}.  It
uses continuous representations of words
\citep{bengio2006neural,mikolov2013distributed} to improve \gls{LDA}
in terms of predictive performance and topic quality.  The \gls{ETM}
defines each topic as a vector on the word embedding space; it then
uses the dot product between each word and the topic embedding to
define the per-topic distribution over words. However, while the
\gls{ETM} better fits large document collections, it cannot analyze a
corpus whose topics shift over time.

In this paper we develop the \gls{DETM}, a model that extends
\gls{DLDA} and the \gls{ETM}. Similarly to \gls{DLDA}, the \gls{DETM}
involves a probabilistic time series to allow the topics to vary
smoothly over time. However, each topic in the \gls{DETM} is a
time-varying vector on the word embedding space. As in the \gls{ETM},
the probability of each word under the \gls{DETM} is a categorical
distribution whose natural parameter depends on the inner product
between the word's embedding and a per-topic embedding representation
of its assigned topic. In contrast to the \gls{ETM}, the topic
embeddings of the \gls{DETM} vary over time.

Given a time-series corpus of documents, we are interested in the
posterior distribution of the topic proportions and the per-time-point
topic embeddings. As for most interesting probabilistic models, the
posterior distribution is intractable to compute; we need to
approximate it.  We use variational inference
\citep{Jordan1999,Blei2017}. To scale up the algorithm to large
datasets, we use data subsampling \citep{Hoffman2013} and amortization
\citep{Gershman2014}; these techniques speed up the learning procedure
and reduce the number of variational parameters.  Additionally, we use
a structured variational approximation parameterized by a \gls{LSTM}
network \citep{Hochreiter1997}.

We use the \gls{DETM} to analyze the transcriptions of the \gls{UN}
general debates from $1970$ to $2015$ \citep{Baturo2017understanding}.
Qualitatively, the \gls{DETM} reveals the topics discussed in the
political debates and their trajectories, which are aligned with
historical events. For example Figure\nobreakspace \ref {fig:topic_evolution_climate_change} 
in Section\nobreakspace \ref {sec:experiments} shows a topic about 
climate change found by the \gls{DETM} that transitions from being mainly 
about the ozone layer in the 1990s to global warming and emissions in $2015$. 

We also used the \gls{DETM} to analyze a dataset of articles from
Science Magazine (1990-1999) and a corpus of ACL abstracts
(1973-2006). We quantitatively assess the \gls{DETM} in terms of
predictive performance and topic quality.  We found that the
\gls{DETM} provides better predictions and topic quality than
\gls{DLDA} in general.

To validate that the gains in performance of the \gls{DETM} is due to
the model and not to the inference procedure used to fit it, we compare to a baseline
that applies the same inference procedure as the \gls{DETM} to
\gls{DLDA}.  We call this baseline \gls{DLDA}-\textsc{rep}. On all
three corpora, we found the \gls{DETM} and \gls{DLDA} both outperform
\gls{DLDA}-\textsc{rep} and that the only advantage of
\gls{DLDA}-\textsc{rep} over \gls{DLDA} is that it is significantly
faster to fit.

The rest of the paper is organized as follows. Section\nobreakspace \ref {sec:related}
discusses the related work.  Section\nobreakspace \ref {sec:background} reviews \gls{LDA},
\gls{DLDA}, and the \gls{ETM}. Section\nobreakspace \ref {sec:model} presents the
\gls{DETM} and the inference algorithm used to fit it. Finally,
Section\nobreakspace \ref {sec:experiments} details the empirical study and
Section\nobreakspace \ref {sec:conclusion} concludes the paper.

 \section{Related Work}
\label{sec:related}

The \gls{DETM} builds on word embeddings, topic models, and dynamic
topic models.

Word embeddings are low-dimensional continuous representations of
words that capture their
semantics~\citep{Rumelhart:1973,Bengio:2003,bengio2006neural,mikolov2013efficient,
  mikolov2013distributed,pennington2014glove,levy2014neural}.  Some
recent work finds embedding representations that vary over time
\citep{Bamler2017,Rudolph2018}. Despite incorporating a time-varying
component, these works have a different goal than the
\gls{DETM}. Rather than modeling the temporal evolution of documents,
they model how the meaning of words shifts over time.  (In future
research, the \gls{DETM} developed here could be used in concert with
these methods.)

There has been a surge of methods that combine word embeddings and
probabilistic topic models.  Some methods modify the prior
distributions over topics in \gls{LDA}~\citep{petterson2010word,
  xie2015incorporating, shi2017jointly, zhao2017word,
  zhao2017metalda}.  These methods use word embeddings as a type of
``side information.''  There are also methods that combine \gls{LDA}
with word embeddings by first converting the discrete text into
continuous observations of embeddings \citep{das2015gaussian,
  xun2016topic, batmanghelich2016nonparametric, xun2017correlated}.
These works adapt \gls{LDA} for real-valued observations, for example
using a Gaussian likelihood.  Still other ways of combining \gls{LDA}
and word embeddings modify the likelihood \citep{nguyen2015improving},
randomly replace words drawn from a topic with the embeddings drawn
from a Gaussian \citep{bunk2018welda}, or use Wasserstein distances to
learn topics and embeddings jointly \citep{xu2018distilled}.  In
contrast to all these methods, the \gls{DETM} uses sequential priors
and is a probabilistic model of discrete data that directly models the
words.

Another line of research improves topic modeling inference through
deep neural networks; these are called neural topic models
\citep{miao2016neural,srivastava2017autoencoding,card2017neural,
  cong2017deep,zhang2018whai}. Most of these works are based on the
variational autoencoder \citep{kingma2014autoencoding} and use
amortized inference \citep{Gershman2014}.  Finally, the \gls{ETM}
\citep{dieng2019topic} is a probabilistic topic model that also makes
use of word embeddings and uses amortization in its inference
procedure.

The first and most common dynamic topic model is \gls{DLDA}
\citep{blei2006dynamic}.  \citet{Bhadury2016} scale up the inference
method of \gls{DLDA} using a sampling procedure.  Other extensions of
\gls{DLDA} use stochastic processes to introduce stronger correlations
in the topic dynamics \citep{Wang2006,Wang2008,Jahnichen2018}.  The
\gls{DETM} is also an extension of \gls{DLDA}, but developed for a
different purpose. The \gls{DETM} better fits the distribution of
words via the use of distributed representations for both the words
and the topics.

 \section{Background}
\label{sec:background}

Here we review the models on which we build the \gls{DETM}. We start
by reviewing \gls{LDA} and the \gls{ETM}; both are non-dynamic topic models.
We then review \gls{DLDA}, the dynamic extension of \gls{LDA}.

Consider a corpus of $D$ documents, where the vocabulary contains $V$ distinct
terms. Let $w_{dn}\in\{1,\ldots,V\}$ denote the $n^\textrm{th}$ word in the
$d^\textrm{th}$ document.

\parhead{\Acrlong{LDA}.} \gls{LDA} is a probabilistic generative model of documents
\citep{blei2003latent}. It considers $K$ topics $\beta_{1:K}$, each of which is a
distribution over the vocabulary. It further considers a vector of topic proportions
$\theta_d$ for each document $d$ in the collection; each element $\theta_{dk}$ 
expresses how prevalent the $k^\textrm{th}$ topic is in that document. In the
generative process of \gls{LDA}, each word is assigned to topic $k$ with probability
$\theta_{dk}$, and the word is then drawn from the distribution $\beta_k$. The
generative process for each document is as follows:
\begin{compactenum}
\item Draw topic proportions $\theta_d \sim \textrm{Dirichlet}(\alpha_{\theta})$.
\item For each word $n$ in the document:
  \begin{compactenum}
    \setlength{\itemindent}{-0.3cm}
  \item Draw topic assignment $z_{dn} \sim \text{Cat}(\theta_d)$.
  \item Draw word $w_{dn} \sim \text{Cat}(\beta_{z_{dn}})$.
  \end{compactenum}
\end{compactenum}
Here, $\textrm{Cat}(\cdot)$ denotes the categorical distribution. \gls{LDA} also places
a Dirichlet prior on the topics, $\beta_k\sim \textrm{Dirichlet}(\alpha_{\beta})$.
The concentration parameters $\alpha_{\beta}$ and $\alpha_{\theta}$ of the
Dirichlet distributions are model hyperparameters.

\parhead{\Acrlong{ETM}.} The \gls{ETM} uses vector representations of words
\citep{Rumelhart:1973,Bengio:2003} to improve the performance of \gls{LDA}
in terms of topic quality and predictive accuracy, specially in the presence
of large vocabularies \citep{dieng2019topic}.
Let $\rho$ be an $L\times V$ matrix containing
$L$-dimensional embeddings of the words in the vocabulary, such that each
column $\rho_v\in\mathbb{R}^L$ corresponds to the embedding representation
of the $v^\textrm{th}$ term. The \gls{ETM} uses the embedding matrix $\rho$
to define each topic $\beta_k$; in particular it sets
\begin{equation}
	\beta_k = \textrm{softmax}(\rho^\top \alpha_k).
\end{equation}
Here, $\alpha_k\in\mathbb{R}^L$ is an embedding representation of the
$k^\textrm{th}$ topic, called \emph{topic embedding}. The topic embedding
is a distributed representation of the topic in the semantic space of
words. The \gls{ETM} uses the topic embeddings in its generative process, 
which is analogous to \gls{LDA}:
\begin{compactenum}
\item Draw topic proportions $\theta_d \sim \mathcal{LN}(0,I)$.
\item For each word $n$ in the document:
  \begin{compactenum}
    \setlength{\itemindent}{-0.3cm}
  \item Draw topic assignment $z_{dn} \sim \text{Cat}(\theta_d)$.
  \item Draw word $w_{dn} \sim \text{Cat}(\textrm{softmax}(\rho^\top \alpha_{z_{dn}}))$.
  \end{compactenum}
\end{compactenum}
The notation $\mathcal{LN}$ in Step 1 refers to the logistic-normal distribution
\citep{Aitchison:1980}, which transforms Gaussian random variables to the simplex.

In using the word representations $\rho_{1:V}$ in the definition of $\beta_{1:K}$, the
\gls{ETM} learns the topics of a corpus in a particular embedding space. 
The intuition behind the \gls{ETM} is that semantically related words
will be assigned to similar topics---since their embedding representations are close,
they will interact similarly with the topic embeddings $\alpha_{1:K}$.

\parhead{\Acrlong{DLDA}.} \gls{DLDA} allows topics to vary over time
to analyze time-series corpora \citep{blei2006dynamic}. The generative
model of \gls{DLDA} differs from \gls{LDA} in that the topics are
time-specific, i.e., they are $\beta_{1:K}^{(t)}$, where
$t\in\{1,\ldots,T\}$ indexes time steps. Moreover, the prior over the
topic proportions $\theta_d$ depends on the time stamp of document
$d$, denoted $t_d\in\{1,\ldots,T\}$. The generative process for each
document is:
\begin{compactenum}
\item Draw topic proportions $\theta_d \sim \mathcal{LN}(\eta_{t_d}, a^2 I)$.
\item For each word $n$ in the document:
  \begin{compactenum}
    \setlength{\itemindent}{-0.3cm}
  \item Draw topic assignment $z_{dn} \sim \text{Cat}(\theta_d)$.
  \item Draw word $w_{dn} \sim \text{Cat}(\beta_{z_{dn}}^{(t_d)})$.
  \end{compactenum}
\end{compactenum}
Here, $a$ is a model hyperparameter and $\eta_{t}$ is a latent variable that controls
the prior mean over the topic proportions at time $t$. To encourage smoothness over
the topics and topic proportions, \gls{DLDA} places random walk priors over
$\beta_{1:K}^{(t)}$ and $\eta_t$,
\begin{align*}
 	 \widetilde{\beta}_k^{(t)}\g\widetilde{\beta}_k^{(t-1)} &\sim\mathcal{N}(\widetilde{\beta}_k^{(t-1)}, \sigma^2 I) \text{ and } 
	 \beta_k^{(t)}=\textrm{softmax}(\widetilde{\beta}_k^{(t)})\\
	\eta_t\g \eta_{t-1} &\sim\mathcal{N}(\eta_{t-1}, \delta^2 I).
\end{align*}
The variables $\widetilde{\beta}_k^{(t)}\in\mathbb{R}^V$ are the transformed topics;
the topics $\beta_k^{(t)}$ are obtained after mapping $\widetilde{\beta}_k^{(t)}$
to the simplex. The hyperparameters $\sigma$ and $\delta$ control the smoothness of
the Markov chains.

\section{The Dynamic Embedded Topic Model}
\label{sec:model}

Here we develop the \gls{DETM}, a model that combines the advantages
of \gls{DLDA} and the \gls{ETM}. Like \gls{DLDA}, it allows the topics
to vary smoothly over time to accommodate datasets that span a large
period of time. Like the \gls{ETM}, the \gls{DETM} uses word
embeddings, allowing it to generalize better than \gls{DLDA} and
improving its topics.  We describe the model in Section\nobreakspace \ref {subsec:model}
and then we develop an efficient structured variational inference algorithm in
Section\nobreakspace \ref {subsec:inference}.

\subsection{Model Description}
\label{subsec:model}

The \gls{DETM} is a dynamic topic model that uses embedding
representations of words and topics.  For each term $v$, it considers
an $L$-dimensional embedding representation $\rho_v$.  The \gls{DETM}
posits an embedding $\alpha_k^{(t)}\in\mathbb{R}^L$ for each topic $k$
at a given time stamp $t=1,\ldots,T$.  That is, the \gls{DETM}
represents each topic as a time-varying real-valued vector, unlike
traditional topic models (where topics are distributions over the
vocabulary). We refer to $\alpha_k^{(t)}$ as \emph{topic embedding}
\citep{dieng2019topic}; it is a distributed representation of the
$k^\mathrm{th}$ topic in the semantic space of words.

The \gls{DETM} forms distributions over the vocabulary using the word
and topic embeddings. Specifically, under the \gls{DETM}, the
probability of a word under a topic is given by the (normalized)
exponentiated inner product between the embedding representation of
the word and the topic's embedding at the corresponding time
step, 
\begin{equation}\label{eq:detm_conditional_lik}
  p(w_{dn}=v\g z_{dn}=k, \alpha_k^{(t_d)})\propto \exp\{ \rho_v^\top \alpha_k^{(t_d)} \}.
\end{equation}
The probability of a particular term is higher when the term's
embedding and the topic's embeddings are in agreement. Therefore,
semantically similar words will be assigned to similar topics, since
their representations are close in the embedding space.

\begin{figure}[t]
	\centering
	\includegraphics[width=0.45\textwidth]{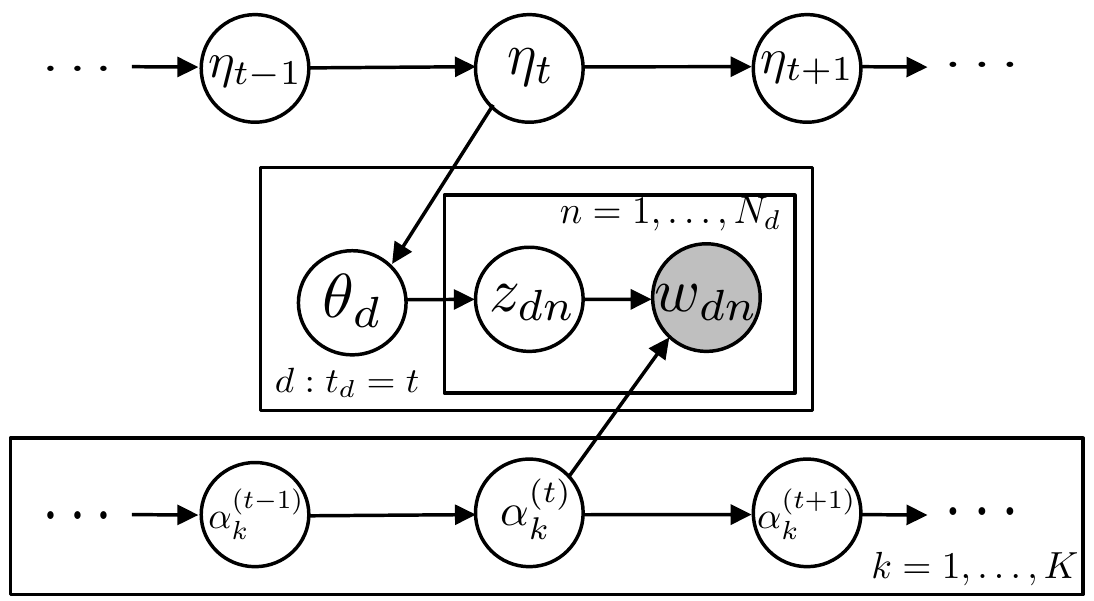}
	\caption{Graphical representation of \gls{DETM}. The topic
          embeddings $\alpha_k^{(t)}$ and the latent means $\eta_t$
          evolve over time. For each document at time step $t$, the
          prior over the topic proportions $\theta_d$ depends on
          $\eta_t$. The variables $z_{dn}$ denote the topic
          assignment; the variables $w_{dn}$ denote the words.}
	\label{fig:graphical_model}
\end{figure}

The \gls{DETM} enforces smooth variations of the topics by using a
Markov chain over the topic embeddings $\alpha_k^{(t)}$. The topic
representations evolve under Gaussian noise with variance
$\gamma^2$,~\looseness=-1
\begin{equation}
	p(\alpha_k^{(t)}\g \alpha_k^{(t-1)})=\mathcal{N}(\alpha_k^{(t-1)}, \gamma^2 I).
\end{equation}
Similarly to \gls{DLDA}, the \gls{DETM} considers time-varying priors
over the topic proportions $\theta_d$. In addition to time-varying
topics, this construction allows the model to capture how the general
topic usage evolves over time.  The prior over $\theta_d$ depends on a
latent variable $\eta_{t_d}$, where recall that $t_d$ is the time
stamp of document $d$,
\begin{align*}
  p(\theta_d\g \eta_{t_d})&=\mathcal{LN}(\eta_{t_d}, a^2 I)  \text{ where } 
  p(\eta_t\g \eta_{t-1})=\mathcal{N}(\eta_{t-1}, \delta^2 I).
\end{align*}
Figure\nobreakspace \ref {fig:graphical_model} depicts the graphical model for the
\gls{DETM}.  The generative process is as follows:
\begin{compactenum}
\item Draw initial topic embedding $\alpha_k^{(0)} \sim \mathcal{N}(0,I)$
\item Draw initial topic proportion mean $\eta_0 \sim \mathcal{N}(0,I)$
\item For time step $t = 1 ,\ldots, T$:
\begin{compactenum}
	\item Draw topic embeddings $\alpha_k^{(t)} \sim \mathcal{N}(\alpha_k^{(t-1)}, \gamma^2 I)$ for $k=1,\ldots,K$
	\item Draw topic proportion means $\eta_t\sim \mathcal{N}(\eta_{t-1}, \delta^2 I)$
 \end{compactenum}
\item For each document $d$:
  \begin{compactenum}
    \setlength{\itemindent}{-0.1cm}
  \item Draw topic proportions $\theta_d \sim \mathcal{LN}(\eta_{t_d}, a^2 I)$.
  \item For each word $n$ in the document:
    \begin{compactenum}
      \setlength{\itemindent}{-0.3cm}
    \item Draw topic assignment $z_{dn} \sim \text{Cat}(\theta_d)$.
    \item Draw word $w_{dn} \sim \text{Cat}(\textrm{softmax}(\rho^\top \alpha_{z_{dn}}^{(t_d)}))$.
    \end{compactenum}
  \end{compactenum}
\end{compactenum}
Steps 1 and 3a give the prior over the topic embeddings; it encourages
smoothness on the resulting topics. Steps 2 and 3b is shared with \gls{DLDA};
it describes the evolution of the prior mean over the topic
proportions. Steps 4a and 4b-i are standard for topic modeling; they
represent documents as distributions over topics and draw a topic
assignment for each word. Step 4b-ii is different---it uses the
$L\times V$ word embedding matrix $\rho$ and the assigned topic
embedding $\alpha_{z_{dn}}^{(t_d)}$ at time instant $t_d$ to form a
categorical distribution over the vocabulary.

Since the \gls{DETM} uses embedding representations of the words, it
learns the topics in a particular embedding space. This aspect of the
model is useful when the embedding of a new word is available, i.e., a
word that does not appear in the corpus. Specifically, consider a term
$v^\star$ that was not seen in the corpus. The \gls{DETM} can assign
it to topics by computing the inner product
$\rho_{v^\star}^\top \alpha_k^{(t)}$, thus leveraging the semantic
information of the word's embedding.

\subsection{Inference Algorithm}
\label{subsec:inference}

We observe a dataset $\mathcal{D}$ of documents
$\{\bw_1,\ldots,\bw_D \}$ and their time stamps $\{t_1,\ldots,t_D\}$.
Fitting a \gls{DETM} involves finding the posterior distribution over
the model's latent variables, $p(\theta,\eta,\alpha\g \mathcal{D})$,
where we have marginalized out the topic assignments $z$ from
Eq.\nobreakspace \ref {eq:detm_conditional_lik} for convenience,\footnote{
  Marginalizing $z_{dn}$ reduces the number of variational parameters
  and avoids discrete latent variables in the inference procedure,
  which is useful to form reparameterization gradients.}
\begin{equation}
	p(w_{dn}\g \alpha_k^{(t_d)}) = \sum_{k=1}^{K} p(w_{dn}\g z_{dn}=k, \alpha_k^{(t_d)}).
\end{equation}
The posterior is intractable.  We approximate it with variational
inference \citep{Jordan1999,Blei2017}.

Variational inference approximates the posterior using a family of
distributions $q_{\nu}(\theta,\eta,\alpha)$. The parameters $\nu$ that
index this family are called variational parameters, and are optimized
to minimize the \gls{KL} divergence between the approximation and the
posterior. Solving this optimization problem is equivalent to
maximizing the \gls{ELBO},
\begin{equation}\label{eq:elbo}
  \Lcal(\nu) = \E{q}{\log p(\mathcal{D},\theta,\eta,\alpha) - \log q_{\nu}(\theta,\eta,\alpha)}.
\end{equation}

To reduce the number of variational parameters and speed-up the
inference algorithm, we use an amortized variational distribution,
i.e., we let the parameters of the approximating distributions be
functions of the data
\citep{Gershman2014,kingma2014autoencoding}. Additionally, we use a
structured variational family to preserve some of the conditional
dependencies of the graphical model \citep{Saul1996}. The specific
variational family in the \gls{DETM} takes the form
\begin{align}\label{eq:q_factorization}
  q(\theta,\eta,\alpha)
  &=
    \prod_d q(\theta_d\g\eta_{t_d},\bw_d)
    \times
    \prod_t q(\eta_t\g \eta_{1:t-1}, \widetilde{\bw}_t)
  \times
    \prod_k \prod_t q(\alpha_k^{(t)}).
\end{align}
(To avoid clutter, we suppress the notation for the variational
parameters.)

The distribution over the topic proportions
$q(\theta_d\g\eta_{t_d},\bw_d)$ is a logistic-normal whose mean and
covariance parameters are functions of both the latent mean
$\eta_{t_d}$ and the bag-of-words representation of document $d$.  In
particular, these functions are parameterized by feed-forward neural
networks that input both $\eta_{t_d}$ and the normalized bag-of-words
representation.  The distribution over the latent means
$q(\eta_t\g \eta_{1:t-1}, \widetilde{\bw}_t)$ depends on all previous
latent means $\eta_{1:t-1}$. We use an \gls{LSTM} to capture this
temporal dependency. We choose a Gaussian distribution
$q(\eta_t\g \eta_{1:t-1}, \widetilde{\bw}_t)$ whose mean and
covariance are given by the output of the \gls{LSTM}.  The input to
the \gls{LSTM} at time $t$ is the average of the bag-of-words
representation of all documents whose time stamp is $t$. Here,
$\widetilde{\bw}_t$ denotes the normalized bag-of-words representation
of all such documents. Finally, unlike \cite{blei2006dynamic}, we do not use structured 
variational inference for the topics. Instead, we use the mean-field family for the
approximation over the topic embeddings, $q(\alpha_k^{(t)})$, for simplicity. 

We optimize the \gls{ELBO} with respect to the variational
parameters. Because the expectations in Eq.\nobreakspace \ref {eq:elbo} are
intractable, we use black box variational inference, obtaining
unbiased gradient estimators with Monte Carlo. In particular, we form
reparameterization gradients
\citep{kingma2014autoencoding,Titsias2014_doubly,rezende2014stochastic}.
To speed up the algorithm, we take a minibatch of documents at each
iteration; this allows to handle large collections of documents
\citep{Hoffman2013}.  We set the learning rate with Adam
\citep{Kingma2015}.  Algorithm\nobreakspace \ref {alg:etm} summarizes the procedure.

\begin{algorithm}[t]
    \caption{Dynamic topic modeling with the \gls{DETM} \label{alg:etm}}
    \SetAlgoNoLine
    \DontPrintSemicolon
    \SetKwInOut{KwInput}{input}
    \SetKwInOut{KwOutput}{output}
    \KwInput{Documents $\{\bw_1,\ldots,\bw_D\}$ and their time stamps $\{t_1,\ldots,t_D\}$}
    Initialize all variational parameters\;
    \For{\emph{iteration} $1,2,3,\ldots$}{
		Sample the latent means and the topic embeddings, $\eta \sim q(\eta\g \widetilde{\bw})$ and $\alpha\sim q(\alpha)$\;
		Compute the topics $\beta_k^{(t)} = \textrm{softmax}(\rho^\top \alpha_k^{(t)})$ for $k=1,\ldots,K$ and $t=1,\ldots,T$\;
		Obtain a minibatch of documents\;
		\For{\emph{each document $d$ in the minibatch}}{
			Sample the topic proportions $\theta_d \sim q(\theta_d\g \eta_{t_d}, \bw_d)$\;
			\For{\emph{each word $n$ in the document}}{
				Compute $p(w_{dn} \g \theta_d) = \sum_k \theta_{dk}\beta_{k,w_{dn}}^{(t_d)}$\;
			}
		}
		Estimate the \acrshort{ELBO} in Eq.\nobreakspace \ref {eq:elbo} and its gradient w.r.t.\ the variational parameters (backpropagation)\;
		Update the model and variational parameters using Adam\;
	}
 \end{algorithm}

\section{Empirical Study}
\label{sec:experiments}

\begin{table*}[t]
	\centering
	\small
	\captionof{table}{Summary statistics of the different datasets under study.}
	\begin{tabular}{cccccc}
	\toprule
	 Dataset & \# Docs Train & \# Docs Val & \# Docs Test & \# Timestamps & Vocabulary \\
	 \midrule
	 \textsc{un}    & $196{,}290$ & $11{,}563$ & $23{,}097$ & $46$ & $12{,}466$ \\
	 \textsc{science}  &  $13{,}894$ &      $819$ &  $1{,}634$ & $10$ & $25{,}987$ \\
	 \textsc{acl}    &   $8{,}936$ &      $527$ &  $1{,}051$ & $31$ & $35{,}108$ \\
	 \bottomrule
	\end{tabular}
	\label{tab:datasets}
\end{table*}

We use the \gls{DETM} to analyze the transcriptions of the \gls{UN}
general debates from $1970$ to $2015$, a corpus of \textsc{acl} abstracts from $1973$ to $2006$, and a set of articles from Science Magazine from $1990$ to $1999$. 
We found the \gls{DETM} provides better predictive power and higher topic quality in general on these datasets when compared to \gls{DLDA}. 

On the transcriptions of the \gls{UN} general debates, we additionally carried out a qualitative analysis of the results. We found that
the \gls{DETM} reveals the temporal evolution of the topics discussed in the debates (such as
climate change, war, poverty, or human rights). 

We compared the \gls{DETM} against two versions of \gls{DLDA}, labeled as \gls{DLDA}
and \gls{DLDA}-\textsc{rep}, which differ only in the inference method (the details are below). 
The comparison of the \gls{DETM} against \gls{DLDA}-\textsc{rep}
reveals that the key to the \gls{DETM}'s performance is the model and not the inference procedure.

\begin{table*}[t]
	\centering
	\small
	\captionof{table}{Predictive performance as measured by held-out perplexity (lower is better) on a document completion task. 
	The \gls{DETM} outperforms both \gls{DLDA} and \gls{DLDA}-\textsc{rep} on all but one corpus. These results also show 
	that the \gls{DETM} gains its advantage through its modeling assumptions and not through its inference procedure.}
	\begin{tabular}{cccc}
	\toprule
	 Method & \textsc{un} & \textsc{science} & \textsc{acl} \\
	 \midrule
	 \gls{DLDA} \citep{blei2006dynamic} & $2393.5$ & $\textbf{3600.7}$ & $4324.2$ \\
	 \gls{DLDA}-\textsc{rep}  & $2931.3$ & $8377.4$ & $5836.7$ \\
	 \acrshort{DETM} & $\textbf{1970.7}$  & $4206.1$ & $\textbf{4120.6}$\\
	 \bottomrule
	\end{tabular}
	\label{tab:quantitative}
\end{table*}

\begin{table*}[t]
	\centering
	\small
	\captionof{table}{Qualitative performance on the \textsc{un} dataset as measured by topic coherence (\textsc{tc}), topic diversity (\textsc{td}), and topic quality (\textsc{tq}). The higher these metrics the better. The \acrshort{DETM} achieves better overall topic quality than \gls{DLDA} and \gls{DLDA}-\textsc{rep}.}
	\begin{tabular}{c|ccc}
	 \midrule
	 Method & \textsc{tc} & \textsc{td} & \textsc{tq}  \\
	 \midrule
	 \gls{DLDA} \citep{blei2006dynamic} & $\textbf{0.1317}$ & $0.6065$ & $0.0799$  \\
	 \gls{DLDA}-\textsc{rep}  & $0.1180$ & $0.2691$ & $0.0318$  \\
	  \acrshort{DETM}  & $0.1206$ & $\textbf{0.6703}$ & $\textbf{0.0809}$  \\
	 \bottomrule
	\end{tabular}
	\label{tab:quantitative_un}
\end{table*}

\begin{table*}[t]
	\centering
	\small
	\captionof{table}{Qualitative performance on the \textsc{science} dataset as measured by topic coherence (\textsc{tc}), topic diversity (\textsc{td}), and topic quality (\textsc{tq}). The higher these metrics the better. The \acrshort{DETM} achieves better overall topic quality than \gls{DLDA} and \gls{DLDA}-\textsc{rep}.}
	\begin{tabular}{c|ccc}
	 \midrule
	 Method & \textsc{tc} & \textsc{td} & \textsc{tq}  \\
	 \midrule
	 \gls{DLDA} \citep{blei2006dynamic} &  $\textbf{0.2392}$ & $0.6502$ & $0.1556$  \\
	 \gls{DLDA}-\textsc{rep}  & $0.0611$ & $0.2290$ & $0.0140$  \\
	  \acrshort{DETM}  & $0.2298$ & $\textbf{0.8215}$ & $\textbf{0.1888}$ \\
	 \bottomrule
	\end{tabular}
	\label{tab:quantitative_sci}
\end{table*}

\begin{table*}[t]
	\centering
	\small
	\captionof{table}{Qualitative performance on the \textsc{acl} dataset as measured by topic coherence (\textsc{tc}), topic diversity (\textsc{td}), and topic quality (\textsc{tq}). The higher these metrics the better. The \acrshort{DETM} achieves better overall topic quality than \gls{DLDA} and \gls{DLDA}-\textsc{rep}.}
	\begin{tabular}{c|ccc}
	 \midrule
	 Method & \textsc{tc} & \textsc{td} & \textsc{tq}  \\
	 \midrule
	 \gls{DLDA} \citep{blei2006dynamic} & $0.1429$ & $0.5904$ & $0.0844$ \\
	 \gls{DLDA}-\textsc{rep}  & $0.1011$ & $0.2589$ & $0.0262$ \\
	  \acrshort{DETM}  & $\textbf{0.1630}$ & $\textbf{0.8286}$ & $\textbf{0.1351}$ \\
	 \bottomrule
	\end{tabular}
	\label{tab:quantitative_acl}
\end{table*}

\parhead{Datasets.}  We study the \gls{DETM} on three datasets.  The
\gls{UN} debates corpus\footnote{Available at
  \url{https://www.kaggle.com/unitednations/un-general-debates}.}
spans $46$ years \citep{Baturo2017understanding}.  Each year, leaders
and other senior officials deliver statements that present their
government's perspective on the major issues in world politics.  The
corpus contains the transcriptions of each country's statement at the
\gls{UN} General Assembly.  We follow \citet{Lefebure2018} and split
the speeches into paragraphs, treating each paragraph as a separate
document.

The second dataset is ten years of \textsc{science} articles, $1990$
to $1999$. The articles are from \textsc{jstor}, an on-line archive of
scholarly journals that scans bound volumes and runs optical character
recognition algorithms on the scans.  This data was used by
\citet{blei2007correlated}.

The third dataset is a collection of articles from $1973$ to $2006$
from the \textsc{acl} Anthology \citep{Bird2008}.  This anthology is a
repository of computational linguistics and natural language
processing papers.

For each dataset, we apply standard preprocessing techniques, such as
tokenization and removal of numbers and punctuation marks.  We also
filter out stop words, i.e., words with document frequency above
$70\%$, as well as standard stop words from a list. Additionally, we
remove low-frequency words, i.e., words that appear in less than a
certain number of documents ($30$ documents for \gls{UN} debates, 
$100$ documents for the \textsc{science} corpus, and $10$ documents for the \textsc{acl} dataset). 
We use $85\%$ randomly chosen documents for training, $10\%$ for testing, and $5\%$ for validation,
and we remove one-word documents from the validation and test
sets. Table\nobreakspace \ref {tab:datasets} summarizes the characteristics of each dataset.

\parhead{Methods.} We compare the \gls{DETM} against two variants of
\gls{DLDA}. One variant is the original model and algorithm of
\citet{blei2006dynamic}. The other variant, which we call \gls{DLDA}-\textsc{rep}, is the 
\gls{DLDA} model of \cite{blei2006dynamic} fitted using mean-field variational inference with the reparameterization trick. 
The comparison against \gls{DLDA}-\textsc{rep} helps us delineate between performance due to the model
and performance due to the inference algorithm.

\begin{figure*}[t]
	\centering
	\includegraphics[width=1.0\linewidth]{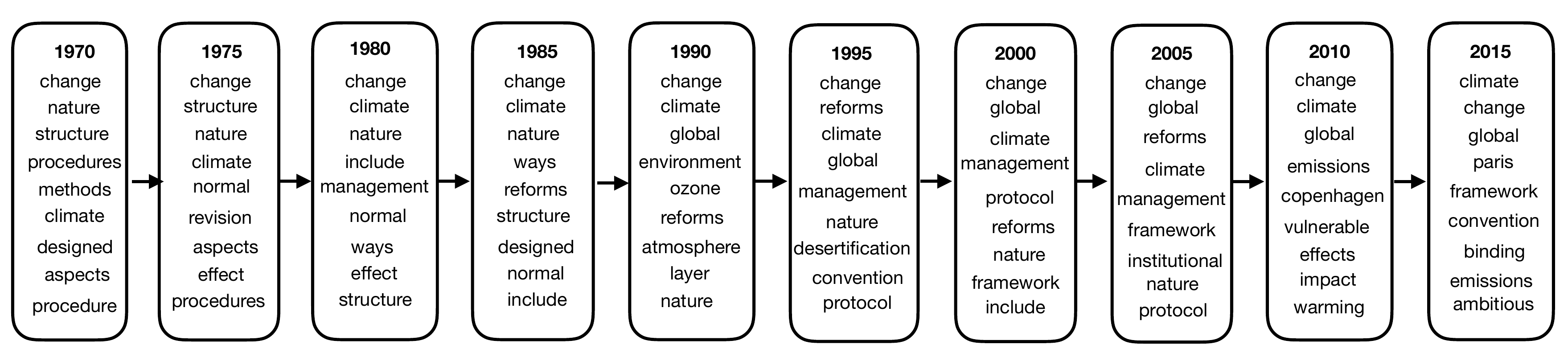}
	\caption{Temporal evolution of the top-$10$ words from a topic about climate change learned by the \acrshort{DETM}. This topic is in agreement with historical events. In the 1990s the destruction of the ozone layer was of major concern. More recently the concern is about global warming. Events such as the Kyoto protocol and the Paris convention are also reflected in this topic's evolution.}
	\label{fig:topic_evolution_climate_change}
	\vspace{-0.2in}
\end{figure*}

\begin{figure*}[!hbpt]
	\centering
	\includegraphics[width=1.0\linewidth]{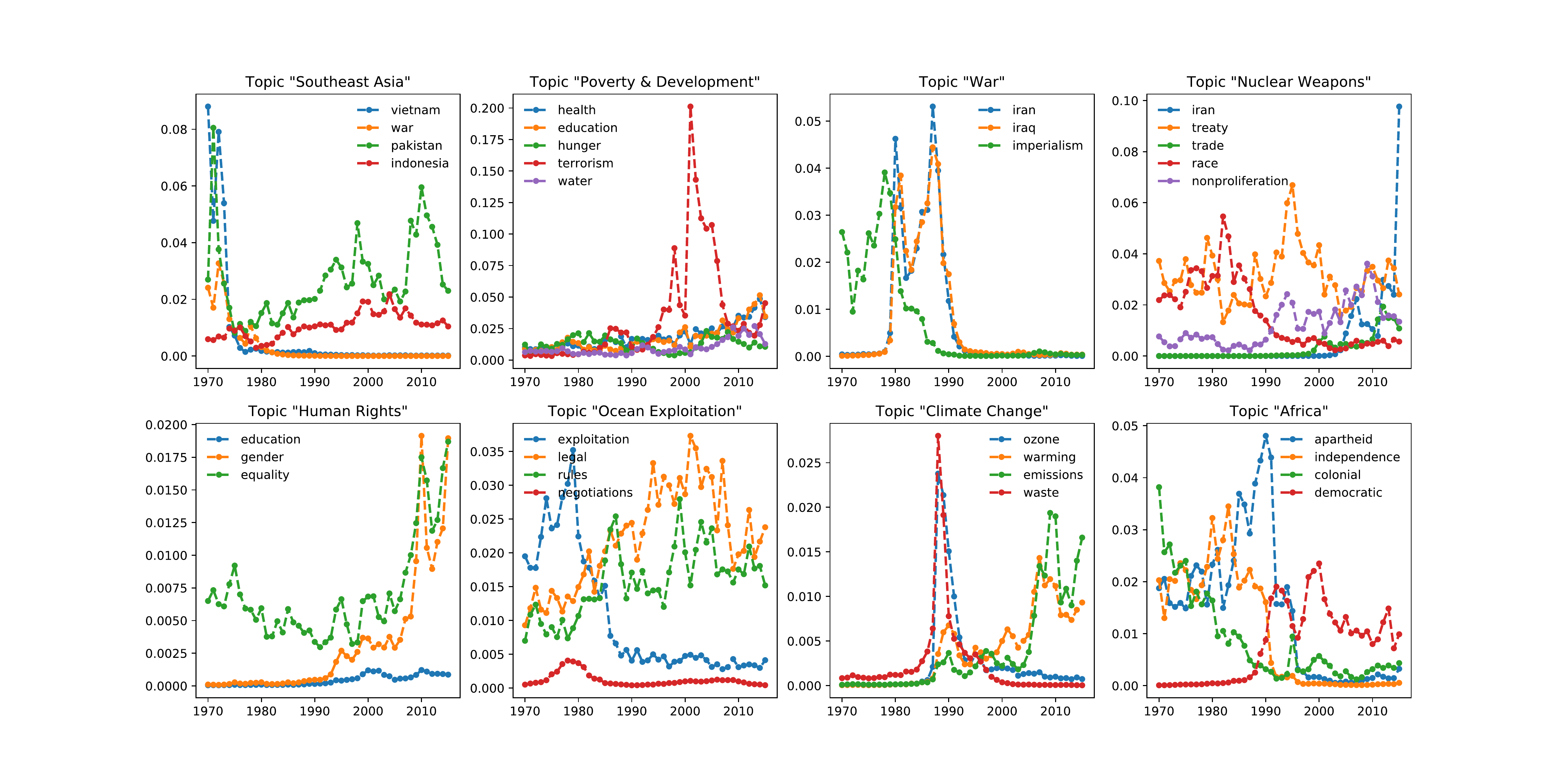}
	\caption{Evolution of word probability across time for eight different topics learned by the \acrshort{DETM}. For each topic, we choose a set of words whose probability shift aligns with historical events (these are not the words with the highest probability in each topic). For example, one interesting finding is the increased relevance of the words ``gender'' and ``equality'' in a topic about human rights.}
	\label{fig:word_evolution}
	\vspace{-0.2in}
\end{figure*}

\parhead{Settings.} We use $50$ topics for all the experiments and
follow \citet{blei2006dynamic} to set the variances of the different priors as
$\delta^2=\sigma^2=\gamma^2=0.005$ and $a^2=1$.

For the \gls{DETM}, we first fit $300$-dimensional word embeddings using skip-gram
\citep{mikolov2013distributed}\footnote{More advanced methods can be used to learn word embeddings. 
We used skip-gram for simplicity and found it leads to good performance.}.
We apply the algorithm in Section\nobreakspace \ref {subsec:inference} using a batch size of $200$
documents for all datasets except for \textsc{acl} for which we used $100$. 
We use a fully connected feed-forward inference network for the topic proportions
$\theta_d$. The network has ReLU activations and $2$ layers of $800$ hidden units each. 
We set the mean and log-variance for $\theta_d$ as linear maps of the output. We applied a small 
dropout rate of $0.1$ to the output of this network before using it to compute the mean and the log-variance. 
For the latent means
$\eta_{1:T}$, each bag-of-word representation $\widetilde{\bw}_t$ is first linearly
mapped to a low-dimensional space of dimensionality $400$. This conforms the input of
an \gls{LSTM} that has $4$ layers of $400$ hidden units
each. The \gls{LSTM} output is then concatenated with the previous latent mean
$\eta_{t-1}$, and the result is linearly mapped to a $K$-dimensional space to get the mean
and log-variance for $\eta_t$.
We apply a small weight decay of $1.2\cdot10^{-6}$ on all network parameters. 
We run Algorithm\nobreakspace \ref {alg:etm} for $1000$ epochs on \textsc{science} and \textsc{acl} 
and for $400$ epochs on the \textsc{un} dataset. The stopping criterion is based on the
held-out log-likelihood on the validation set. The learning rate is set to $0.001$ for 
the \textsc{un} and \textsc{science} datasets and to $0.0008$ on the \textsc{acl} corpus. 
We fixed the learning rate throughout training. We clip the norm of the gradients of the \gls{ELBO}
to $2.0$ to stabilize training. 

We fit \gls{DLDA} using the published code of \citet{blei2006dynamic}. (See \url{https://github.com/blei-lab/dtm}.) 
To fit \gls{DLDA}, \citet{blei2006dynamic} derived a bound of the \gls{ELBO} to enable
a coordinate-ascent inference algorithm that also uses Kalman filtering and smoothing as a 
subroutine. Besides loosening the variational bound on the log-marginal likelihood of the data,
this algorithm presents scalability issues both in terms of the number of topics and in terms of the vocabulary. 
For example fitting \gls{DLDA} took almost two days on each dataset whereas we only required 
less than $6$ hours for the \gls{DETM}. 

To fit \gls{DLDA}-\textsc{rep} we leverage recent advances in variational inference to overcome these issues. 
We use stochastic optimization based on reparameterization gradients and we draw 
batches of $1{,}000$ documents at each iteration. We collapse the discrete latent topic 
indicators $z_{dn}$ to enable the reparameterization gradients, and we use a fully factorized 
Gaussian approximation for the rest of the latent variables, except for $\eta_{1:T}$, for which we use a
full-covariance Gaussian for each of its dimensions. We initialize \gls{DLDA} using
\gls{LDA}. In particular, we run $5$ epochs of \gls{LDA} followed by $120$ epochs of
\gls{DLDA}. For \gls{DLDA}, we use RMSProp \citep{Tieleman2012} to set the step size,
setting the learning rate to $0.05$ for the mean parameters and to $0.005$ for the
variance parameters.

\parhead{Quantitative results.}
We compare the \gls{DETM}, \gls{DLDA}, and \gls{DLDA}-\textsc{rep} according to two metrics: perplexity on a
document completion task and topic quality. The perplexity is obtained by computing the
probability of each word in the second half of a test document, conditioned on the
first half \citep{rosenzvi2004author,wallach2009evaluation}. To obtain the topic quality,
we combine two metrics. The first metric is topic coherence; it provides a quantitative
measure of the interpretability of a topic \citep{mimno2011optimizing}. We obtain the
topic coherence by taking the average pointwise mutual information of two words drawn randomly
from the same document \citep{lau2014machine}; this requires to approximate word
probabilities with empirical counts. The second metric is topic diversity; it is the
percentage of unique words in the top $25$ words of all topics \citep{dieng2019topic}.
Diversity close to $0$ indicates redundant topics. We obtain both topic coherence and topic
diversity by averaging over time. Finally, topic quality is defined as
the product between topic coherence and diversity \citep{dieng2019topic}.~\looseness=-1

Table\nobreakspace \ref {tab:quantitative}, Table\nobreakspace \ref {tab:quantitative_un}, Table\nobreakspace \ref {tab:quantitative_sci}, and Table\nobreakspace \ref {tab:quantitative_acl} show 
that the \gls{DETM} outperforms both \gls{DLDA} and \gls{DLDA}-\textsc{rep} according to both perplexity and topic quality 
on almost all datasets. In particular, the \gls{DETM} finds more diverse and coherent topics. 
We posit this is due to its use of embeddings. 

\parhead{Qualitative results.}
The \gls{DETM} finds that the topics' evolution over time are in agreement with historical
events. As an example, Figure\nobreakspace \ref {fig:topic_evolution_climate_change} shows the trajectory of 
a topic on climate change. In the 1990s, protecting the ozone layer was the primary
concern; more recently the topic has shifted towards global warming and reducing 
the greenhouse gas emissions. Some events on
climate change, such as the Kyoto protocol (1997) or the Paris convention (2016),
are also reflected in the topic's evolution.

We now examine the evolution of the probability of individual words. 
Figure\nobreakspace \ref {fig:word_evolution} shows these probabilities for a variety
of words and topics. For example, the probability of the word ``Vietnam'' in a topic on Southeast
Asia decays after the end of the war in 1975. In a topic about nuclear weapons, the
concern about the arms ``race'' between the USA and the Soviet Union eventually decays,
and ``Iran'' becomes more relevant in recent years. Similarly, words like ``equality''
and ``gender'' become more important in recent years within a topic about human rights.
Note that the names of the topics are subjective; we assigned the names inspired by the top
words in each topic (the words in Figure\nobreakspace \ref {fig:word_evolution} are not necessarily the most likely words within
each topic). One example is the topic on climate change, whose top words are shown in Figure\nobreakspace \ref {fig:topic_evolution_climate_change}. Another example is the topic on human rights, which
exhibits the words ``human'' and ``rights'' consistently at the top across all time steps.

 \section{Conclusion}
\label{sec:conclusion}

We developed the \gls{DETM}, a probabilistic model of documents that combines word embeddings and dynamic latent Dirichlet allocation (\gls{DLDA}). 
The \gls{DETM} models each word with a categorical distribution parameterized by the dot product between the embedding of the word and 
an embedding representation of its assigned topic. Each topic embedding is a time-varying vector in the embedding
space of words. Using a random walk prior over these topic embeddings, the \gls{DETM} uncovers smooth topic trajectories. 
We applied the \gls{DETM} to analyze three different corpora and found that the \gls{DETM} outperforms \gls{DLDA} both in terms of 
predictive performance and topic quality while requiring significantly less time to fit.

\subsection*{Acknowledgements}
This work is funded by ONR N00014-17-1-2131, NIH 1U01MH115727-01, DARPA SD2 FA8750-
18-C-0130, ONR N00014-15-1-2209, NSF CCF-1740833, the Alfred P. Sloan Foundation, 2Sigma,
Amazon, and NVIDIA. Francisco J.\ R.\ Ruiz is supported by the European Union's Horizon 2020 research and innovation programme under the Marie Sk\l{}odowska-Curie grant agreement No.\ 706760. Adji B.\ Dieng is supported by a Google PhD Fellowship.

\bibliographystyle{apa}
\bibliography{main}

\begin{thebibliography}{}

\bibitem[\protect\astroncite{Aitchison and Shen}{1980}]{Aitchison:1980}
Aitchison, J. and Shen, S. (1980).
\newblock Logistic normal distributions: some properties and uses.
\newblock {\em Biometrika}, 67(2):261--272.

\bibitem[\protect\astroncite{Bamler and Mandt}{2017}]{Bamler2017}
Bamler, R. and Mandt, S. (2017).
\newblock Dynamic word embeddings.
\newblock In {\em International Conference on Machine Learning}.

\bibitem[\protect\astroncite{Batmanghelich
  et~al.}{2016}]{batmanghelich2016nonparametric}
Batmanghelich, K., Saeedi, A., Narasimhan, K., and Gershman, S. (2016).
\newblock Nonparametric spherical topic modeling with word embeddings.
\newblock In {\em Association for Computational Linguistics}.

\bibitem[\protect\astroncite{Baturo et~al.}{2017}]{Baturo2017understanding}
Baturo, A., Dasandi, N., and Mikhaylov, S. (2017).
\newblock Understanding state preferences with text as data: introducing the
  {UN} general debate corpus.
\newblock {\em Research \& Politics}, 4:1--9.

\bibitem[\protect\astroncite{Bengio et~al.}{2003}]{Bengio:2003}
Bengio, Y., Ducharme, R., Vincent, P., and Janvin, C. (2003).
\newblock A neural probabilistic language model.
\newblock {\em Journal of Machine Learning Research}, 3:1137--1155.

\bibitem[\protect\astroncite{Bengio et~al.}{2006}]{bengio2006neural}
Bengio, Y., Schwenk, H., Sen{\'e}cal, J.-S., Morin, F., and Gauvain, J.-L.
  (2006).
\newblock Neural probabilistic language models.
\newblock In {\em Innovations in Machine Learning}, pages 137--186. Springer.

\bibitem[\protect\astroncite{Bhadury et~al.}{2016}]{Bhadury2016}
Bhadury, A., Chen, J., Zhu, J., and Liu, S. (2016).
\newblock Scaling up dynamic topic models.
\newblock In {\em International World Wide Web Conference}.

\bibitem[\protect\astroncite{Bird et~al.}{2008}]{Bird2008}
Bird, S., Dale, R., Dorr, B., Gibson, B., Joseph, M., Kan, M.-Y., Lee, D.,
  Powley, B., Radev, D., and Tan, Y.~F. (2008).
\newblock The {ACL} anthology reference corpus: a reference dataset for
  bibliographic research in computational linguistics.
\newblock In {\em International Conference on Language Resources and
  Evaluation}.

\bibitem[\protect\astroncite{Blei}{2012}]{blei2012probabilistic}
Blei, D.~M. (2012).
\newblock Probabilistic topic models.
\newblock {\em Communications of the ACM}, 55(4):77--84.

\bibitem[\protect\astroncite{Blei et~al.}{2017}]{Blei2017}
Blei, D.~M., Kucukelbir, A., and McAuliffe, J.~D. (2017).
\newblock Variational inference: {a} review for statisticians.
\newblock {\em Journal of the American Statistical Association},
  112(518):859--877.

\bibitem[\protect\astroncite{Blei and Lafferty}{2006}]{blei2006dynamic}
Blei, D.~M. and Lafferty, J.~D. (2006).
\newblock Dynamic topic models.
\newblock In {\em International Conference on Machine Learning}.

\bibitem[\protect\astroncite{Blei and Lafferty}{2007}]{blei2007correlated}
Blei, D.~M. and Lafferty, J.~D. (2007).
\newblock A correlated topic model of {S}cience.
\newblock {\em The Annals of Applied Statistics}, 1(1):17--35.

\bibitem[\protect\astroncite{Blei and McAuliffe}{2007}]{blei2007supervised}
Blei, D.~M. and McAuliffe, J.~D. (2007).
\newblock Supervised topic models.
\newblock In {\em Advances in Neural Information Processing Systems}.

\bibitem[\protect\astroncite{Blei et~al.}{2003}]{blei2003latent}
Blei, D.~M., Ng, A.~Y., and Jordan, M.~I. (2003).
\newblock Latent {D}irichlet allocation.
\newblock {\em Journal of Machine Learning Research}, 3:993--1022.

\bibitem[\protect\astroncite{Boyd-Graber
  et~al.}{2017}]{boydgraber2017applications}
Boyd-Graber, J., Hu, Y., and Mimno, D. (2017).
\newblock Applications of topic models.
\newblock {\em Foundations and Trends in Information Retrieval},
  11(2--3):143--296.

\bibitem[\protect\astroncite{Bunk and Krestel}{2018}]{bunk2018welda}
Bunk, S. and Krestel, R. (2018).
\newblock {WELDA}: enhancing topic models by incorporating local word context.
\newblock In {\em Proceedings of the 18th ACM/IEEE on Joint Conference on
  Digital Libraries}, pages 293--302. ACM.

\bibitem[\protect\astroncite{Card et~al.}{2017}]{card2017neural}
Card, D., Tan, C., and Smith, N.~A. (2017).
\newblock A neural framework for generalized topic models.
\newblock In {\em arXiv:1705.09296}.

\bibitem[\protect\astroncite{Cong et~al.}{2017}]{cong2017deep}
Cong, Y., Chen, B., Liu, H., and Zhou, M. (2017).
\newblock Deep latent {D}irichlet allocation with topic-layer-adaptive
  stochastic gradient {R}iemannian {MCMC}.
\newblock In {\em International Conference on Machine Learning}.

\bibitem[\protect\astroncite{Das et~al.}{2015}]{das2015gaussian}
Das, R., Zaheer, M., and Dyer, C. (2015).
\newblock Gaussian {LDA} for topic models with word embeddings.
\newblock In {\em Association for Computational Linguistics and International
  Joint Conference on Natural Language Processing (Volume 1: Long Papers)}.

\bibitem[\protect\astroncite{Dieng et~al.}{2019}]{dieng2019topic}
Dieng, A.~B., Ruiz, F.~J., and Blei, D.~M. (2019).
\newblock Topic modeling in embedding spaces.
\newblock {\em arXiv preprint arXiv:1907.04907}.

\bibitem[\protect\astroncite{Gershman and Goodman}{2014}]{Gershman2014}
Gershman, S.~J. and Goodman, N.~D. (2014).
\newblock Amortized inference in probabilistic reasoning.
\newblock In {\em Annual Meeting of the Cognitive Science Society}.

\bibitem[\protect\astroncite{Hochreiter and Schmidhuber}{1997}]{Hochreiter1997}
Hochreiter, S. and Schmidhuber, J. (1997).
\newblock Long short-term memory.
\newblock {\em Neural Computation}, 9(8):1735--1780.

\bibitem[\protect\astroncite{Hoffman et~al.}{2013}]{Hoffman2013}
Hoffman, M.~D., Blei, D.~M., Wang, C., and Paisley, J. (2013).
\newblock Stochastic variational inference.
\newblock {\em Journal of Machine Learning Research}, 14:1303--1347.

\bibitem[\protect\astroncite{J\"{a}hnichen et~al.}{2018}]{Jahnichen2018}
J\"{a}hnichen, P., Wenzel, F., Kloft, M., and Mandt, S. (2018).
\newblock Scalable generalized dynamic topic models.
\newblock In {\em Artificial Intelligence and Statistics}.

\bibitem[\protect\astroncite{Jordan et~al.}{1999}]{Jordan1999}
Jordan, M.~I., Ghahramani, Z., Jaakkola, T.~S., and Saul, L.~K. (1999).
\newblock An introduction to variational methods for graphical models.
\newblock {\em Machine Learning}, 37(2):183--233.

\bibitem[\protect\astroncite{Kingma and Ba}{2015}]{Kingma2015}
Kingma, D.~P. and Ba, J.~L. (2015).
\newblock Adam: a method for stochastic optimization.
\newblock In {\em International Conference on Learning Representations}.

\bibitem[\protect\astroncite{Kingma and Welling}{2014}]{kingma2014autoencoding}
Kingma, D.~P. and Welling, M. (2014).
\newblock Auto-encoding variational {B}ayes.
\newblock In {\em International Conference on Learning Representations}.

\bibitem[\protect\astroncite{Lafferty and Blei}{2005}]{Lafferty2005correlated}
Lafferty, J.~D. and Blei, D.~M. (2005).
\newblock Correlated topic models.
\newblock In {\em Advances in Neural Information Processing Systems}.

\bibitem[\protect\astroncite{Lau et~al.}{2014}]{lau2014machine}
Lau, J.~H., Newman, D., and Baldwin, T. (2014).
\newblock Machine reading tea leaves: automatically evaluating topic coherence
  and topic model quality.
\newblock In {\em Conference of the European Chapter of the Association for
  Computational Linguistics}.

\bibitem[\protect\astroncite{Lefebure}{2018}]{Lefebure2018}
Lefebure, L. (2018).
\newblock Exploring the {UN} general debates with dynamic topic models.
\newblock Available online at \url{https://towardsdatascience.com}.

\bibitem[\protect\astroncite{Levy and Goldberg}{2014}]{levy2014neural}
Levy, O. and Goldberg, Y. (2014).
\newblock Neural word embedding as implicit matrix factorization.
\newblock In {\em Neural Information Processing Systems}, pages 2177--2185.

\bibitem[\protect\astroncite{Miao et~al.}{2016}]{miao2016neural}
Miao, Y., Yu, L., and Blunsom, P. (2016).
\newblock Neural variational inference for text processing.
\newblock In {\em International Conference on Machine Learning}.

\bibitem[\protect\astroncite{Mikolov et~al.}{2013a}]{mikolov2013efficient}
Mikolov, T., Chen, K., Corrado, G., and Dean, J. (2013a).
\newblock Efficient estimation of word representations in vector space.
\newblock {\em ICLR Workshop Proceedings. arXiv:1301.3781}.

\bibitem[\protect\astroncite{Mikolov et~al.}{2013b}]{mikolov2013distributed}
Mikolov, T., Sutskever, I., Chen, K., Corrado, G.~S., and Dean, J. (2013b).
\newblock Distributed representations of words and phrases and their
  compositionality.
\newblock In {\em Neural Information Processing Systems}.

\bibitem[\protect\astroncite{Mimno et~al.}{2009}]{Mimno2009polylingual}
Mimno, D., Wallach, H.~M., Naradowsky, J., Smith, D.~A., and McCallum, A.
  (2009).
\newblock Polylingual topic models.
\newblock In {\em Conference on Empirical Methods in Natural Language
  Processing}.

\bibitem[\protect\astroncite{Mimno et~al.}{2011}]{mimno2011optimizing}
Mimno, D., Wallach, H.~M., Talley, E., Leenders, M., and McCallum, A. (2011).
\newblock Optimizing semantic coherence in topic models.
\newblock In {\em Conference on Empirical Methods in Natural Language
  Processing}.

\bibitem[\protect\astroncite{Nguyen et~al.}{2015}]{nguyen2015improving}
Nguyen, D.~Q., Billingsley, R., Du, L., and Johnson, M. (2015).
\newblock Improving topic models with latent feature word representations.
\newblock {\em Transactions of the Association for Computational Linguistics},
  3:299--313.

\bibitem[\protect\astroncite{Pennington et~al.}{2014}]{pennington2014glove}
Pennington, J., Socher, R., and Manning, C.~D. (2014).
\newblock {GloVe}: global vectors for word representation.
\newblock In {\em Conference on Empirical Methods on Natural Language
  Processing}.

\bibitem[\protect\astroncite{Petterson et~al.}{2010}]{petterson2010word}
Petterson, J., Buntine, W., Narayanamurthy, S.~M., Caetano, T.~S., and Smola,
  A.~J. (2010).
\newblock Word features for latent {D}irichlet allocation.
\newblock In {\em Advances in Neural Information Processing Systems}, pages
  1921--1929.

\bibitem[\protect\astroncite{Rezende et~al.}{2014}]{rezende2014stochastic}
Rezende, D.~J., Mohamed, S., and Wierstra, D. (2014).
\newblock Stochastic backpropagation and approximate inference in deep
  generative models.
\newblock {\em arXiv preprint arXiv:1401.4082}.

\bibitem[\protect\astroncite{Rosen-Zvi et~al.}{2004}]{rosenzvi2004author}
Rosen-Zvi, M., Griffiths, T., Steyvers, M., and Smyth, P. (2004).
\newblock The author-topic model for authors and documents.
\newblock In {\em Uncertainty in Artificial Intelligence}.

\bibitem[\protect\astroncite{Rudolph and Blei}{2018}]{Rudolph2018}
Rudolph, M. and Blei, D.~M. (2018).
\newblock Dynamic embeddings for language evolution.
\newblock In {\em International World Wide Web Conference}.

\bibitem[\protect\astroncite{Rumelhart and Abrahamson}{1973}]{Rumelhart:1973}
Rumelhart, D. and Abrahamson, A. (1973).
\newblock A model for analogical reasoning.
\newblock {\em Cognitive Psychology}, 5(1):1--28.

\bibitem[\protect\astroncite{Saul and Jordan}{1996}]{Saul1996}
Saul, L.~K. and Jordan, M.~I. (1996).
\newblock Exploiting tractable substructures in intractable networks.
\newblock In {\em Advances in Neural Information Processing Systems}.

\bibitem[\protect\astroncite{Shi et~al.}{2017}]{shi2017jointly}
Shi, B., Lam, W., Jameel, S., Schockaert, S., and Lai, K.~P. (2017).
\newblock Jointly learning word embeddings and latent topics.
\newblock In {\em ACM SIGIR Conference on Research and Development in
  Information Retrieval}.

\bibitem[\protect\astroncite{Srivastava and
  Sutton}{2017}]{srivastava2017autoencoding}
Srivastava, A. and Sutton, C. (2017).
\newblock Autoencoding variational inference for topic models.
\newblock {\em arXiv preprint arXiv:1703.01488}.

\bibitem[\protect\astroncite{Tieleman and Hinton}{2012}]{Tieleman2012}
Tieleman, T. and Hinton, G. (2012).
\newblock Lecture 6.5-{RMSPROP}: divide the gradient by a running average of
  its recent magnitude.
\newblock Coursera: Neural Networks for Machine Learning, 4.

\bibitem[\protect\astroncite{Titsias and
  L\'{a}zaro-Gredilla}{2014}]{Titsias2014_doubly}
Titsias, M.~K. and L\'{a}zaro-Gredilla, M. (2014).
\newblock Doubly stochastic variational {B}ayes for non-conjugate inference.
\newblock In {\em International Conference on Machine Learning}.

\bibitem[\protect\astroncite{Wallach et~al.}{2009}]{wallach2009evaluation}
Wallach, H.~M., Murray, I., Salakhutdinov, R., and Mimno, D. (2009).
\newblock Evaluation methods for topic models.
\newblock In {\em International Conference on Machine Learning}.

\bibitem[\protect\astroncite{Wang et~al.}{2008}]{Wang2008}
Wang, C., Blei, D.~M., and Heckerman, D. (2008).
\newblock Continuous time dynamic topic models.
\newblock In {\em Uncertainty in Artificial Intelligence}.

\bibitem[\protect\astroncite{Wang and McCallum}{2006}]{Wang2006}
Wang, X. and McCallum, A. (2006).
\newblock Topics over time: a non-{M}arkov continuous-time model of topical
  trends.
\newblock In {\em ACM SIGKDD}.

\bibitem[\protect\astroncite{Xie et~al.}{2015}]{xie2015incorporating}
Xie, P., Yang, D., and Xing, E. (2015).
\newblock Incorporating word correlation knowledge into topic modeling.
\newblock In {\em Conference of the North American chapter of the Association
  for Computational Linguistics: Human Language Technologies}.

\bibitem[\protect\astroncite{Xu et~al.}{2018}]{xu2018distilled}
Xu, H., Wang, W., Liu, W., and Carin, L. (2018).
\newblock Distilled {W}asserstein learning for word embedding and topic
  modeling.
\newblock In {\em Advances in Neural Information Processing Systems}.

\bibitem[\protect\astroncite{Xun et~al.}{2016}]{xun2016topic}
Xun, G., Gopalakrishnan, V., Ma, F., Li, Y., Gao, J., and Zhang, A. (2016).
\newblock Topic discovery for short texts using word embeddings.
\newblock In {\em International Conference on Data Mining}.

\bibitem[\protect\astroncite{Xun et~al.}{2017}]{xun2017correlated}
Xun, G., Li, Y., Zhao, W.~X., Gao, J., and Zhang, A. (2017).
\newblock A correlated topic model using word embeddings.
\newblock In {\em IJCAI}, pages 4207--4213.

\bibitem[\protect\astroncite{Zhang et~al.}{2018}]{zhang2018whai}
Zhang, H., Chen, B., Guo, D., and Zhou, M. (2018).
\newblock {WHAI}: {W}eibull hybrid autoencoding inference for deep topic
  modeling.
\newblock In {\em International Conference on Learning Representations}.

\bibitem[\protect\astroncite{Zhao et~al.}{2017a}]{zhao2017word}
Zhao, H., Du, L., and Buntine, W. (2017a).
\newblock A word embeddings informed focused topic model.
\newblock In {\em Asian Conference on Machine Learning}.

\bibitem[\protect\astroncite{Zhao et~al.}{2017b}]{zhao2017metalda}
Zhao, H., Du, L., Buntine, W., and Liu, G. (2017b).
\newblock Meta{LDA}: A topic model that efficiently incorporates meta
  information.
\newblock In {\em International Conference on Data Mining}.

\end{thebibliography}

\end{document}